\documentclass[10pt,twocolumn,letterpaper]{article}

\usepackage[pagenumbers]{cvpr} 

\usepackage{times}
\usepackage{epsfig}
\usepackage{graphicx}
\usepackage{amsmath}
\usepackage{amssymb}
\usepackage{algorithm}
\usepackage{algorithmic}

\usepackage[dvipsnames,svgnames,x11names]{xcolor}
\definecolor{citecolor}{RGB}{119,185,0} 
\usepackage[pagebackref=false,breaklinks=true,letterpaper=true,colorlinks,citecolor=citecolor,bookmarks=false]{hyperref}
\usepackage{colortbl}

\newlength\savewidth\newcommand\shline{\noalign{\global\savewidth\arrayrulewidth
  \global\arrayrulewidth 1pt}\hline\noalign{\global\arrayrulewidth\savewidth}}
  
\begin{document}

\title{PiPa: Pixel- and Patch-wise Self-supervised Learning \\ for Domain Adaptative Semantic Segmentation}

\author{
Mu Chen$^1$, Zhedong Zheng$^2$, Yi Yang$^3$, Tat-Seng Chua$^2$\\
$^1$ReLER, AAII, University of Technology Sydney \\$^2$Sea-NExT Joint Lab, School of 
Computing, National University of Singapore \\ $^3$Zhejiang University}

\maketitle

\begin{abstract}
   Unsupervised Domain Adaptation (UDA) aims to enhance the generalization of the learned model to other domains. The domain-invariant knowledge is transferred from the model trained on labeled source domain, \eg, video game, to unlabeled target domains, \eg, real-world scenarios, saving annotation expenses. Existing UDA methods for semantic segmentation usually focus on minimizing the inter-domain discrepancy of various levels, \eg, pixels, features, and predictions, for extracting domain-invariant knowledge. 
   However, the primary intra-domain knowledge, such as context correlation inside an image, remains under-explored. 
   In an attempt to fill this gap, we propose a unified pixel- and patch-wise self-supervised learning framework, called \textbf{PiPa}, for domain adaptive semantic segmentation that facilitates intra-image pixel-wise correlations and patch-wise semantic consistency against different contexts.
   \begin{figure}[]
  \centering
  \includegraphics[width=0.95\linewidth]{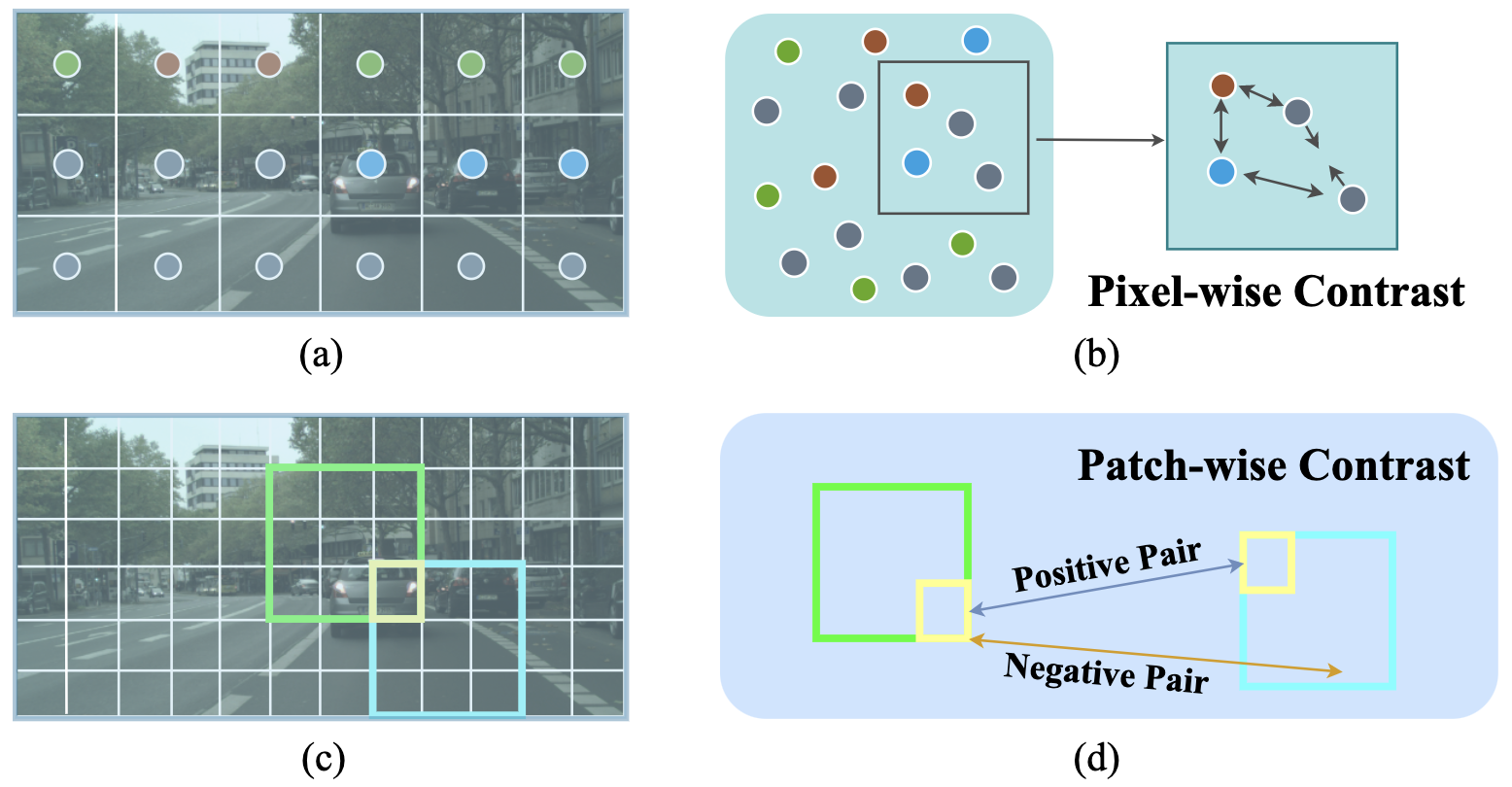}
  \caption{Different from existing works, we focus on mining the intra-domain knowledge, and argue that the contextual structure between pixels and patches can facilitate the model learning the domain-invariant knowledge in a self-supervised manner.
  In particular, our proposed training framework: 
  (1) motivates intra-class compactness and inter-class dispersion by pulling closer the pixel-wise intra-class features and pushing away inter-class features within the image (see a\&b at the top row);
  and  (2) maintains the local patch consistency against different contexts, such as the \textcolor{yellow}{yellow} local patch in the \textcolor{green}{green} and the \textcolor{blue}{blue} patch (see the bottom row c\&d). 
  }
  \vspace{-1em}
  \label{fig1}
\end{figure}
   The proposed framework exploits the inherent structures of intra-domain images, which: (1) explicitly encourages learning the  discriminative pixel-wise features with intra-class compactness and inter-class separability, and (2) motivates the robust feature learning of the identical patch against different contexts or fluctuations.
   Extensive experiments verify the effectiveness of the proposed method, which obtains competitive accuracy on the two widely-used UDA benchmarks, \ie, 75.6 mIoU on GTA$\rightarrow$Cityscapes and 68.2 mIoU on Synthia$\rightarrow$Cityscapes. Moreover, our method is compatible with other UDA approaches to further improve the performance without introducing extra parameters. 
  
\end{abstract}

\section{Introduction}

Prevailing models, \eg, Convolutional Neural Networks (CNNs)~\cite{long2015fully, chen2017deeplab} and Visual Transformers~\cite{zheng2021rethinking, liu2021swin}, have achieved significant progress in the semantic segmentation task, but such networks are data-hungry, which usually require large-scale training datasets with pixel-level annotations. The annotation prerequisites are hard to meet in  real-world scenarios. 
To address the shortage in the training data, one straightforward idea is to 
access the abundant synthetic data and the corresponding pixel-level annotations generated by computer graphics \cite{StephanRRichter2016PlayingFD, GermanRos2016TheSD}. However, there exist domain gaps between synthetic images and real-world images in terms of
illumination, weather, and camera hyper-parameters~\cite{ZuxuanWu2019ACEAT,wang2022multiple,fang2022behavioral}. 
To minimize such a gap, researchers resort to unsupervised domain adaptation (UDA) to transfer the knowledge from labeled source-domain data to the unlabeled target-domain environment. 

The key idea underpinning UDA is to learn the shared domain-invariant knowledge. 
One line of works, therefore, investigates techniques to mitigate the discrepancy of data distribution between the source domain and the target domain at different levels, such as pixel level \cite{hoffman2018cycada, li2019bidirectional,yang2020fda, ZuxuanWu2019ACEAT}, feature level \cite{luo2019significance,huang2018domain}, and prediction level \cite{tsai2018learning, tsai2019domain, FeiPan2020UnsupervisedIA,vu2019advent}. 
These inter-domain alignment approaches have 
achieved significant performance improvement compared to the basic source-only methods, but usually overlook the intra-domain knowledge. 

Another potential paradigm to address the lack of training data is self-supervised learning, which mines the visual knowledge from unlabeled data. 
One common optimization objective is to learn invariant representation against various augmentations, such as rotation~\cite{komodakis2018unsupervised}, colorization~\cite{zhang2016colorful}, mixup~\cite{sun2022self} and random erasing~\cite{zhong2020random}. 
Within this context, some prior UDA works~\cite{zheng2019unsupervised,zheng2021rectifying} have explored self-supervised learning to mine the domain-invariant knowledge. However, a common issue behind these methods is that their pipeline is
relatively simple that only considers the prediction consistency against dropout or different network depths. 
Therefore, the optimization target is not well aligned with the local context-focused semantic segmentation task, which restricts the self-supervised gain from data.

In the light of the above observation, we propose a learning strategy that jointly couples the inter-domain and intra-domain learning in a unified network. In particular, we introduce a unified Pixel- and Patch-wise self-supervised learning framework (\textbf{PiPa}) to enhance the intra-domain knowledge mining without extra annotations. 
As the name implies, PiPa explores the pixel-to-pixel and patch-to-patch relation for regularizing the segmentation feature space. Our approach is based on two implicit priors: 
(1) 
the feature of the same-class pixels should be kept consistent with the category prototype; and 
(2) 
the feature within a patch should maintain robustness against different contexts. 
As shown in Figure~\ref{fig1}, image pixels are mapped into an embedding space (Figure~\ref{fig1} (b) and (d)). 
For the \textbf{pixel-wise contrast}, we explicitly facilitate discriminative feature learning by pulling pixel embeddings of the same category closer while pushing those of different categories away (Figure~\ref{fig1} 
(b)). 
Considering the \textbf{patch-wise contrast}, we randomly crop two image patches with an overlapping region (the yellow region in Figure~\ref{fig1} (c) and (d)) from an unlabeled image. 
The overlapping region of the two patches should not lose its spatial information and maintain the prediction consistency even against two different contexts. 
The proposed method is orthogonal to other existing domain-alignment works. We re-implement two competitive baselines, and show that our framework consistently improves the segmentation accuracy over other existing works.

Our contributions are as follows: 
(1) Different from existing works on inter-domain alignment, we focus on mining domain-invariant knowledge from the original domain in a self-supervised manner. We propose a unified Pixel- and Patch-wise self-supervised learning framework to harness both pixel- and patch-wise consistency against different contexts, which is well-aligned with the segmentation task.
(2) Our self-supervised learning method does not require extra annotations, and is compatible with other existing UDA frameworks. The effectiveness of PiPa has been tested by extensive ablation studies, and it achieves competitive accuracy on two commonly used UDA benchmarks, namely 75.6 mIoU on GTA$\rightarrow$Cityscapes and 68.2 mIoU on Synthia$\rightarrow$Cityscapes.

\section{Related Work}
\subsection{Unsupervised Domain Adaptation}
Unsupervised domain adaptation (UDA) enables a model to transfer scalable knowledge from a label-rich source domain to a label-scarce target domain, which improves the adaptation performance. 
Some pioneering works \cite{hoffman2018cycada, wu2018dcan} propose to transfer the visual style of the source domain data to the target domain using CycleGAN \cite{zhu2017unpaired}, which allows the model to be trained on labeled data with target style. Later UDA methods can mainly be grouped into two categories according to the technical routes: adversarial training \cite{tsai2018learning, yang2020fda, FeiPan2020UnsupervisedIA, luo2019taking, luo2021category, vu2019advent} and self-training \cite{YangZou2018UnsupervisedDA, zou2019confidence, KeMei2020InstanceAS, kang2020adversarial, PanZhang2021PrototypicalPL, WilhelmTranheden2020DACSDA, QianyuZhou2022UncertaintyAwareCR}. Adversarial training methods aim to learn domain-invariant knowledge based on adversarial domain alignment. For instance, Tsai \etal \cite{tsai2018learning} and Luo \etal \cite{luo2019taking} learn domain-invariant representations based on a min-max adversarial optimization game, where a feature extractor is trained to fool a domain discriminator and thus helps to obtain aligned feature distributions. 
However, as shown in \cite{zheng2022adaptive}, unstable adversarial training methods usually lead to suboptimal performance.
Another line of work harnesses self-training to create pseudo labels for the target domain data using the model trained by labeled source domain data. Then the model is re-trained by pseudo labels. Pseudo labels can be pre-computed either offline \cite{YangZou2018UnsupervisedDA, yang2020fda} or generated online \cite{WilhelmTranheden2020DACSDA, hoyer2022daformer}. Due to considerable discrepancies in data distributions between two domains, pseudo labels inevitably contains noise. To decrease the influence of faculty labels, Zou \etal \cite{YangZou2018UnsupervisedDA, zou2019confidence} adopts pseudo labels with high confidence. Taking one step further, Zheng \etal \cite{zheng2019unsupervised} conducts the domain alignment to create reliable pseudo labels. Furthermore, some variants leverage specialized sampling \cite{KeMei2020InstanceAS} and uncertainty \cite{zheng2021rectifying} to learn from the noisy  pseudo labels. 
In addition to the two mainstream practices mentioned above, researchers also conducted extensive attempts such as entropy minimization \cite{vu2019advent, chen2019domain}, image translation \cite{ShaohuaGuo2021LabelFreeRC, JinyuYang2020LabelDrivenRF}, and combining adversarial training and self-training \cite{li2019bidirectional, HaoranWang2020ClassesMA,zheng2022adaptive}. Recently, Chen \etal \cite{chen2022deliberated} establish a deliberated domain bridging (DDB) that allows the source and target domain to align and interact in the intermediate space.
Different from the above-mentioned works, we focus on further mining the domain-invariant knowledge in a self-supervised manner. We harness the pixel- and patch-wise contrast, which is well aligned with the local context-focused semantic segmentation task. The proposed method is orthogonal with the above-mentioned approaches, and thus is complementary with existing ones to further boost the result.  



\subsection{Contrastive Learning}
Contrastive learning is one of the most prominent unsupervised representation learning methods~\cite{AaronvandenOord2018RepresentationLW, ZhirongWu2018UnsupervisedFL, XinleiChen2020ImprovedBW, TingChen2020ASF, KaimingHe2022MomentumCF}, which contrasts similar (positive) data pairs against dissimilar (negative) pairs, thus learning discriminative feature representations.
For instance, Wu \etal~\cite{ZhirongWu2018UnsupervisedFL} learn feature representations at the instance level and store the instance class representation vector in a memory bank. He~\etal~\cite{KaimingHe2022MomentumCF} match encoded features to a dynamic dictionary which is updated with a momentum strategy. Chen~\etal \cite{TingChen2020ASF} propose to engender negative samples from large mini-batches without any memory bank. In the domain adaptative image classification task, contrastive learning is utilized to align feature space of different domains \cite{GuoliangKang2019ContrastiveAN, SaeidMotiian2017UnifiedDS}. 

A few recent studies utilize contrastive learning to improve the performance of semantic segmentation task \cite{XinlongWang2020DenseCL, WouterVanGansbeke2021UnsupervisedSS, WenguanWang2021ExploringCP, liu2021domain, BinhuiXie2022SePiCoSP, jiang2022prototypical}. For example, Wang \etal \cite{XinlongWang2020DenseCL} have designed and optimized a self-supervised learning framework for better visual pre-training. Gansbeke \etal \cite{WouterVanGansbeke2021UnsupervisedSS} applies contrastive learning between features from different saliency masks in an unsupervised setting. Recently, Huang \etal \cite{huang2022category} tackles UDA by considering instance contrastive learning as a dictionary look-up operation, allowing learning of category-discriminative feature representations. Xie \etal \cite{xie2021spcl} presents a semantic prototype-based contrastive learning method for fine-grained class alignment. Other works explore contrastive learning either in a fully supervised manner \cite{WenguanWang2021ExploringCP, BinhuiXie2022SePiCoSP} or in a semi-supervised manner \cite{InigoAlonso2021SemiSupervisedSS, zhou2022domain, lai2021semi}. Most methods above either target image-wise instance separation or tend to learn pixel correspondence alone. 
In this work, we argue that pixel- or patch-wise contrast can be formulated in a similar format but with different effect regions. 
Different from existing works, we explicitly take both pixel- and patch-wise contrast into consideration which is well aligned with segmentation domain adaptation. 
The multi-grained self-supervised learning on both pixel and patch are complementary to each other, and can mine the domain-invariant context feature.


\begin{figure*}
  \centering
    \includegraphics[width=0.9\linewidth]{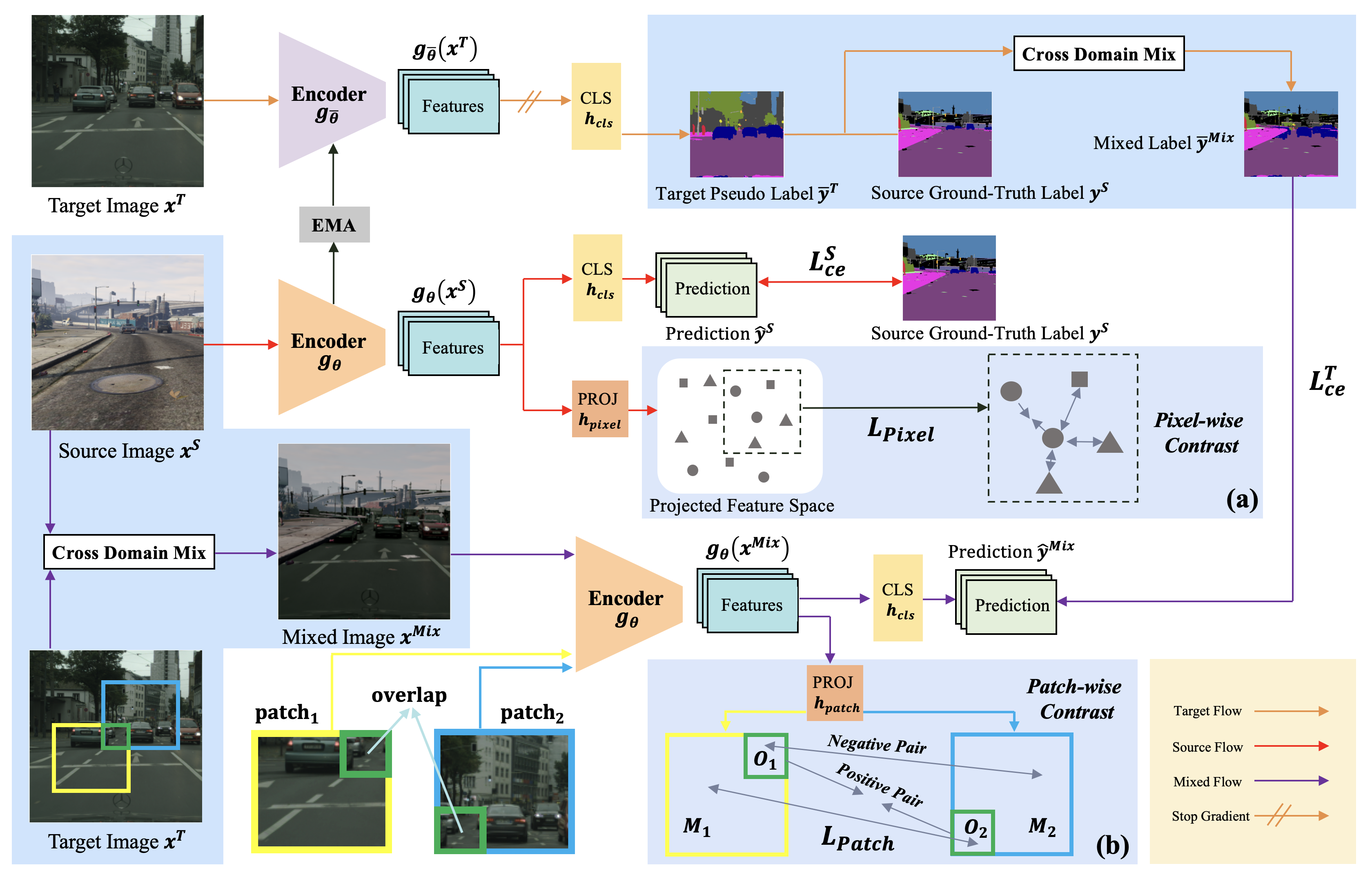}
    \vspace{-.1in}
    \caption{A brief illustration of our Pixel- and Patch-wise self-supervised learning Framework (PiPa). 
    Given the labeled source data $\left\{\left(x^{S}, y^{S}\right)\right\}$, we calculate the segmentation prediction $\hat{y}^S$ with the backbone $g_\theta$ and the classification head $h_{cls}$, supervised by the basic segmentation loss $L_{ce}^S$.  During training, we leverage the moving averaged model $\boldsymbol{g}_{\bar{\theta}}$ to estimate the pseudo label $\bar{y}^T$ to craft the mixed label $\bar{y}^{Mix}$ based on the category.  According to the mixed label, we copy the corresponding regions as the mixed data $x^{Mix}$. We also deploy the model $g_\theta$ and the head $h_{cls}$ to obtain the mixed prediction $\hat{y}^{Mix}$ supervised by $L_{ce}^T$. 
    Except for the above-mentioned basic segmentation losses, we propose Pixel-wise Contrast and Patch-wise Contrast. 
    In \textbf{(a)}, we regularize the pixel embedding space by computing pixel-to-pixel contrast: impelling positive-pair embeddings closer, and pushing away the negative embeddings. In \textbf{(b)}, we regularize the patch-wise consistency between projected patch $\mathbf{O}_1$ and $\mathbf{O}_2$. Similarly, we harness the patch-wise contrast, which pulls positive pair, \ie, two features at the same location of $\mathbf{O}_1$ and $\mathbf{O}_2$ closer, while pushing negative pairs apart, \ie, any two features in $\mathbf{M}_1 \cup \mathbf{M}_2$ at different locations. During inference, we drop the two projection heads $h_{patch}$ and $h_{pixel}$ and only keep $g_\theta$ and $h_{cls}$. }
    \label{fig2}
\end{figure*}

\section{Methods}
We first introduce the problem definition and conventional segmentation losses for semantic segmentation domain adaptation. 
Then we shed light on the proposed component of our framework PiPa, \ie, Pixel-wise Contrast and Patch-wise Contrast, both of which work on local regions to mine the inherent contextual structures. We finally also raise a discussion on the mechanism of the proposed method.


\noindent\textbf{Problem Definition.} 
As shown in Figure~\ref{fig2}, given the source-domain synthetic data $X^S=\left\{x^S_u\right\}_{u=1}^U$ labeled by $Y^S=\left\{y_u^{S}\right\}_{u=1}^{U}$ and the unlabelled target-domain real-world data $X^T=\left\{x^T_v\right\}_{v=1}^V$ 
, where $U$ and $V$ are the numbers of images in the source and target domain, respectively. 
The label $Y^S$ belongs to $C$ categories.
Domain adaptive semantic segmentation intends to learn a mapping function that projects the input data $X^T$ to the segmentation prediction $Y^T$ in the target domain. 

\noindent\textbf{Basic Segmentation Losses.}  Similar to existing works~\cite{zheng2019unsupervised,zou2019confidence}, we learn the basic source-domain knowledge by adopting the segmentation loss on the source domain, which can be formulated as:
\begin{equation}
\label{eq:1}
\mathcal{L}_{ce}^S=\mathbb{E}\left[-p_u^S \log h_{cls}(g_\theta\left(x_u^S \right))\right],
\end{equation}
where $p_u^S$ is the 
one-hot vector of the label $y_u^S$, and the value $p_u^S(c)$ equals to 1 if $c==y_u^S$ otherwise 0. 
We harness the visual backbone $g_\theta$, and 2-layer multilayer perceptrons (MLPs) $h_{cls}$ for segmentation category prediction.

To mine the knowledge from the target domain, we generate pseudo labels $\bar{Y}^T = \{\bar{y_v}^T\}$  for the target domain data $X^T$ by a teacher network $g_{\bar{\theta}}$ during training \cite{WilhelmTranheden2020DACSDA, zhou2022context}, where $\bar{y_v}^T = argmax(h_{cls} g_{\bar{\theta}}(x^T_v))$. 
In practice, the teacher network $g_{\bar{\theta}}$ is set as the exponential moving average of the weights of the student network $g_\theta$ after each training iteration \cite{tarvainen2017mean,zheng2022adaptive}. 
Considering that there are no labels for the target-domain data, the network $g_\theta$ is trained on the pseudo label $\bar{y_v}^T$ generated by the teacher model $g_{\bar{\theta}}$. Therefore, the segmentation loss can be formulated as:
\begin{equation}
\label{eq:2}
\mathcal{L}_{ce}^T=\mathbb{E}\left[-\bar{p_v}^T \log h_{cls}( g_\theta\left(x_v^T\right))\right],
\end{equation}
where $\bar{p_v}^T$ is the one-hot vector of the pseudo label $\bar{y_v}^T$.
We observe that pseudo labels inevitably introduce noise considering the data distribution discrepancy between two domains. Therefore, we set a threshold that only the pixels whose prediction confidence is higher than the threshold are accounted for the loss.
In practice, we also follow \cite{WilhelmTranheden2020DACSDA, hoyer2022daformer} to mix images from both domains to facilitate stable training. Specifically, the label $\bar{y}^{Mix}$ is generated by copying the random 50\% categories in $y^S$ and pasting such class areas to the target-domain pseudo label $\bar{y}^T$. Similarly, we also paste the corresponding pixel area in $x^S$ to the target-domain input $x^T$ as $x^{Mix}$.
Therefore, the target-domain segmentation loss is updated as:  
\begin{equation}
\label{eq:3}
\mathcal{L}_{ce}^T=\mathbb{E}\left[-\bar{p_v}^{Mix} \log  h_{cls}(g_\theta\left(x_v^{Mix}\right))\right],
\end{equation}
where $\bar{p_v}^{Mix}$ is the probability vector of the mixed label $\bar{y_v}^{Mix}$. Since we deploy the copy-and-paste strategy instead of the conventional mixup~\cite{zhang2017mixup}, the mixed labels are still one-hot.

\noindent\textbf{Pixel-wise Contrast.} 
We note that the above-mentioned segmentation loss does not explicitly consider the inherent context within the image, which is crucial to the local-focused segmentation task. Therefore, we study the feasibility of self-supervised learning in mining intra-domain knowledge.  We propose pixel-wise contrast to encourage the network to learn the 
correlation between 
labeled pixels.
Given the labels of each pixel $y^S$, we regard image pixels of the same class $C$ as positive samples and the rest pixels in $x^S$ belonging to the other classes are the negative samples. The pixel-wise contrastive loss can be derived as:
\begin{equation}
\label{eq:4}
\mathcal{L}_{\text{Pixel}} =- \sum_{C(i) = C(j)} \log \frac{r\left(e_i, e_j\right)}{\sum_{k=1}^{N_{pixel}} r\left(e_i, e_k\right)},
\end{equation}
where 
$e$ is the feature map extracted by the projection head $e = h_{pixel}g_{\theta}(x)$, and $N_{pixel}$ is the number of pixels. $e_i$ denotes the $i$-th feature on the feature map $e$.
$r$ denotes the similarity between the two pixel features. In particular, we deploy 
the exponential cosine similarity $r\left(e_i, e_j\right)=\exp \left(s\left(e_i,e_j\right)/ \tau\right)$, where $s$ is cosine similarity between two pixel features $e_i$ and $e_j$, and $\tau$ is the temperature. 
As shown in Figure~\ref{fig2}, with the guide of pixel-wise contrastive loss, the pixel embeddings of the same class are pulled close and those of the other classes are pushed apart, which promotes intra-class compactness and inter-class separability. 
It is worth noting that we collect negative samples not only from the current image but also from images within the current training batch.
The choice of training samples is crucial for contrastive learning. In this work, 
we regard the negative pairs with high similarity scores and the positive pairs with low similarity as hard samples. Practically, we select the top $10 \%$ nearest negatives and $10 \%$ farthest positives to aggregate the loss.

\noindent\textbf{Patch-wise Contrast.} 
Except for the pixel-wise contrast and segmentation loss, we also introduce patch-wise contrast to enhance the consistency within a local patch. 
In particular, given unlabeled target image $x^T$, we leverage the network $g_\theta$ to extract the feature map of two partially overlapping patches. 
The cropped examples are shown at the bottom of Figure~\ref{fig2}. 
We deploy an independent head $h_{patch}$ with 2-layer MLPs to further project the output feature maps to the embedding space for comparison. 
We denote the overlapping-region as $\mathbf{O}_1$ and $\mathbf{O}_2$ from $\mathbf{M}_1$ and $\mathbf{M}_2$ respectively. 
We argue that the output features of the overlapping region should be invariant to the contexts. Therefore, 
we encourage that 
each feature in $\mathbf{O}_1$ to be consistent with the corresponding feature of the same location in $\mathbf{O}_2$. 
Similar to pixel-wise contrast, as shown in Figure~\ref{fig2} module (b), we regard two features at the same position of $\mathbf{O}_1$ and $\mathbf{O}_2$ as positive pair, 
and any two features in $\mathbf{M}_1$ and $\mathbf{M}_2$ at different positions of the original image are treated as a negative pair. 
Given a target-domain input $x^T$, the patch-wise contrast loss can be formulated as:
\begin{equation}
\label{eq:5}
    \mathcal{L}_{\text {Patch}}=- \sum_{O_1(i)= O_2(j)} \log \frac{r\left({f}_{i}, f_{j}\right)}{\sum_{k=1}^{N_{patch}} r\left({f}_{i}, f_k\right)},
\end{equation}
where $f$ is the feature map extracted by the projection head $f = h_{patch}g_{\theta}(x)$, and $N_{patch}$ is the number of pixels in $M_1 \cup M_2$. $f_i$ denotes $i$-th feature in the map. If $O_1(i)= O_2(j)$, $f_j$ denote the corresponding points of $f_i$ in the context of $O_2$.  
Similarly, $r$ denotes the exponential function of the cosine similarity as the one in pixel contrast. 
It is worth noting that we also enlarge the negative sample pool. In practice, the rest feature $f_k$ not only comes from the union set $M_1 \cup M_2$, but also from other training images within the current batch.

\noindent\textbf{Total Loss.} The overall training objective is the combination of pixel-level cross-entropy loss and the proposed PiPa: 
\begin{equation}
\label{eq:6}
\mathcal{L}_{total}=\mathcal{L}_{\mathrm{ce}}^S+\mathcal{L}_{\mathrm{ce}}^T+\alpha \mathcal{L}_{\text {Pixel}}+\beta \mathcal{L}_{\text {Patch}},
\end{equation}
where 
$\alpha$ and $\beta$ are the weights for pixel-wise contrast $\mathcal{L}_{\text{Pixel}}$ and patch-wise contrast $\mathcal{L}_{\text{Patch}}$, respectively.  We summarize the pipeline of PiPa in Algorithm \ref{alg}.

\noindent\textbf{Discussion.} \textbf{1. Correlation between Pixel and Patch Contrast.} Both pixel and patch contrast are derived from instance-level contrastive learning and share a common underlying idea, \ie, contrast, but they work at different effect regions, \ie, pixel-wise and patch-wise. 
The pixel contrast explores the pixel-to-pixel category correlation over the whole image, while patch-wise contrast imposes regularization on the semantic patches from a local perspective. 
Therefore, the two kinds of contrast are complementary and can work in a unified way to mine the intra-domain inherent context within the data.
\textbf{2. What is the advantage of the proposed framework?} Traditional UDA methods focus on learning shared inter-domain knowledge. Differently, we are motivated by the objectives of UDA semantic segmentation in a bottom-up manner, and thus leverage rich pixel correlations in the training data to facilitate intra-domain knowledge learning. 
By explicitly regularizing the feature space via PiPa, we enable the model to explore the inherent intra-domain context in a self-supervised setting, \ie, pixel-wise and patch-wise,  without extra parameters or annotations. 
Therefore, PiPa could be effortlessly incorporated into existing UDA approaches to achieve better results without extra overhead during testing.
\textbf{3. Difference from conventional contrastive learning.}
Conventional contrastive learning methods typically tend to perform contrast in the instance or pixel level alone \cite{ZhirongWu2018UnsupervisedFL, huang2022category, WenguanWang2021ExploringCP}. 
We formulate pixel- and patch-wise contrast in a similar format but focus on the local effect regions within the images, which is well aligned with the local-focused segmentation task. 
We show that the proposed local contrast, \ie, pixel- and patch-wise contrasts, regularizes the domain adaptation training and guides the model to shed more light on the intra-domain context. 
Our experiment also verifies this point that pixel- and patch-wise contrast facilitates smooth edges between different categories and yields a higher accuracy on small objects.

\begin{algorithm}[t]
	\renewcommand{\algorithmicrequire}{\textbf{Input:}}
	\renewcommand{\algorithmicensure}{\textbf{Output:}}
	\caption{PiPa algorithm}
	\label{alg}
	\begin{algorithmic}[1]
 	\REQUIRE Source-domain data $X^S$, Source-domain labels $Y^S$, Target domain data $X^T$, segmentation network that contains segmentation encoder $g_\theta$, classification head $h_{\text{cls}}$, pixel projection head $h_{\text{pixel}}$, patch projection head $h_{\text{patch}}$, the total iteration number T$_{\text{total}}$. \\
		\STATE Initialize network parameter $\theta$ with ImageNet pre-trained parameters. Initialize teacher network $\bar{\theta}$ randomly
		\FOR{iteration = 1 to T$_{\text{total}}$}
		\STATE $x^S,\,y^S \sim U.$
        \STATE $x^T \sim V.$
        \STATE $\bar{y}^T \leftarrow argmax\left(h_{cls}\left(\bar{g_\theta}\left(x^T\right)\right)\right).$
        \STATE $x^{Mix},\,\bar{y}^{Mix} \gets$ Augmentation and pseudo label from mixing $x^S, y^S, x^T$ and $\bar{y}^T.$
        \STATE Compute predictions \\ $\hat{y}^S \leftarrow
        argmax\left(h_{cls}\left(g_\theta\left(x^S\right)\right)\right),$ \\ $\hat{y}^{Mix} \leftarrow argmax\left(h_{cls}\left(g_\theta\left(x^{Mix}\right)\right)\right).$
        \STATE Compute loss for the mini-batch    $\mathcal{L}_{\text {total}}.$
        \STATE Compute $\nabla_\theta \mathcal{L}_{\text{total}}$ by backpropagation.
        \STATE Perform stochastic gradient descent. 
        \STATE Update teacher network $\bar{\theta}$ with $\theta.$
        \ENDFOR
        \RETURN $g_{\theta}$ and $h_{cls}.$
	\end{algorithmic}  
\end{algorithm}

\section{Experiment}

\begin{table*}[!t]
	\centering
	\caption{Quantitative comparison with previous UDA methods on GTA $\rightarrow$ Cityscapes. We present pre-class IoU and mIoU. The best accuracy in every column is in \textbf{bold}.}
	\label{table:gtacity}
	\resizebox{\linewidth}{!}{
	\begin{tabular}{c|ccccccccccccccccccc|c}
		\shline
		Method & Road & SW & Build & Wall & Fence & Pole & TL & TS & Veg. & Terrain & Sky & PR & Rider & Car & Truck & Bus & Train & Motor & Bike & mIoU\\
		\shline
		\hline
		AdaptSegNet~\cite{tsai2018learning} & 86.5 & 36.0 & 79.9& 23.4 & 23.3 & 23.9 & 35.2 & 14.8 & 83.4 & 33.3 & 75.6 & 58.5 & 27.6 & 73.7 & 32.5 & 35.4 & 3.9 & 30.1 & 28.1 & 42.4\\ 
		CyCADA \cite{hoffman2018cycada} & 86.7 & 35.6 & 80.1 & 19.8 & 17.5 & 38.0 & 39.9 & 41.5 & 82.7 & 27.9 & 73.6 & 64.9 & 19.0 & 65.0 & 12.0 & 28.6 & 4.5 & 31.1 & 42.0 & 42.7 \\
		CLAN~\cite{luo2019taking} & 87.0 & 27.1 & 79.6 & 27.3 & 23.3 & 28.3 & 35.5 & 24.2 & 83.6 & 27.4 & 74.2 & 58.6 & 28.0 & 76.2 & 33.1 & 36.7 & 6.7 & 31.9 & 31.4 & 43.2 \\
		MaxSquare~\cite{chen2019domain} & 88.1 & 27.7 & 80.8 & 28.7 & 19.8 & 24.9 & 34.0 & 17.8 & 83.6 & 34.7 & 76.0 & 58.6 & 28.6 & 84.1 & 37.8 & 43.1 & 7.2 & 32.3 & 34.2 & 44.3 \\
		ASA~\cite{zhou2020affinity} & 89.2 & 27.8 & 81.3 & 25.3 & 22.7 & 28.7 & 36.5 & 19.6 & 83.8 & 31.4 & 77.1 & 59.2 & 29.8 & 84.3 & 33.2 & 45.6 & 16.9 & 34.5 & 30.8 & 45.1 \\
		AdvEnt~\cite{vu2019advent} & 89.4 & 33.1 & 81.0 & 26.6 & 26.8 & 27.2 & 33.5 & 24.7 & 83.9 & 36.7 & 78.8 & 58.7 & 30.5 & 84.8 & 38.5 & 44.5 & 1.7 & 31.6 & 32.4 & 45.5 \\
		MRNet~\cite{zheng2019unsupervised}  & 89.1 & 23.9 & 82.2 & 19.5 & 20.1 & 33.5 & 42.2 & 39.1 & 85.3 & 33.7 & 76.4 & 60.2 & 33.7 & 86.0 & 36.1 & 43.3 & 5.9 & 22.8 & 30.8 & 45.5 \\
		APODA~\cite{yang2020adversarial}  & 85.6 & 32.8 & 79.0 & 29.5 & 25.5 & 26.8 & 34.6 & 19.9 & 83.7 & 40.6 & 77.9 & 59.2 & 28.3 & 84.6 & 34.6 & 49.2 & 8.0 & 32.6 & 39.6 & 45.9 \\
		CBST \cite{YangZou2018UnsupervisedDA} & 91.8 & 53.5 & 80.5 & 32.7 & 21.0 & 34.0 & 28.9 & 20.4 & 83.9 & 34.2 & 80.9 & 53.1 & 24.0 & 82.7 & 30.3 & 35.9 & 16.0 & 25.9 & 42.8 & 45.9\\
		PatchAlign~\cite{tsai2019domain} & 92.3 & 51.9 & 82.1 & 29.2 & 25.1 & 24.5 & 33.8 & 33.0 & 82.4 & 32.8 & 82.2 & 58.6 & 27.2 & 84.3 & 33.4 & 46.3 & 2.2 & 29.5 & 32.3 & 46.5 \\
		MRKLD \cite{zou2019confidence} & 91.0 & 55.4 & 80.0 & 33.7 & 21.4 & 37.3 & 32.9 & 24.5 & 85.0 & 34.1 & 80.8 & 57.7 & 24.6 & 84.1 & 27.8 & 30.1 & 26.9 & 26.0 & 42.3 & 47.1\\
		BL~\cite{li2019bidirectional} & 91.0 & 44.7 & 84.2 & 34.6 & 27.6 & 30.2 & 36.0 & 36.0 & 85.0 & 43.6 & 83.0 & 58.6 & 31.6 & 83.3 & 35.3 & 49.7 & 3.3 & 28.8 & 35.6 & 48.5 \\
		DT~\cite{wang2020differential} & 90.6 & 44.7 & 84.8 & 34.3 & 28.7 & 31.6 & 35.0 & 37.6& 84.7 & 43.3 & 85.3 & 57.0 & 31.5 & 83.8 & 42.6 & 48.5 & 1.9 & 30.4 & 39.0 & 49.2\\  

        FADA~\cite{HaoranWang2020ClassesMA} & 91.0 & 50.6 & 86.0 & 43.4 & 29.8 & 36.8 & 43.4 & 25.0 & 86.8 & 38.3 & 87.4 & 64.0 & 38.0 & 85.2 & 31.6 & 46.1 & 6.5 & 25.4 & 37.1 & 50.1\\ 
		Uncertainty~\cite{zheng2021rectifying} & 90.4 & 31.2 & 85.1 & 36.9 & 25.6 & 37.5 & 48.8 & 48.5 & 85.3 & 34.8 & 81.1 & 64.4 & 36.8 & 86.3 & 34.9 & 52.2 & 1.7 & 29.0 & 44.6 & 50.3 \\
		FDA~\cite{yang2020fda} & 92.5 & 53.3 & 82.4 & 26.5 & 27.6 & 36.4 & 40.6 & 38.9 & 82.3 & 39.8 & 78.0 & 62.6 & 34.4 & 84.9 & 34.1 & 53.1 & 16.9 & 27.7 & 46.4 & 50.5\\
		Adaboost~\cite{zheng2022adaptive} & 90.7 & 35.9 & 85.7 & 40.1 & 27.8 & 39.0 & 49.0 & 48.4 & 85.9 & 35.1 & 85.1 & 63.1 & 34.4 & 86.8 & 38.3 & 49.5 & 0.2 & 26.5 & 45.3 & 50.9\\
		DACS \cite{WilhelmTranheden2020DACSDA} & 89.9 & 39.7 & 87.9 & 30.7 & 39.5 & 38.5 & 46.4 & 52.8 & 88.0 & 44.0 & 88.8 & 67.2 & 35.8 & 84.5 & 45.7 & 50.2 & 0.0 & 27.3 & 34.0 & 52.1\\
		CorDA \cite{wang2021domain} & 94.7 & 63.1 & 87.6 & 30.7 & 40.6 & 40.2 & 47.8 & 51.6 & 87.6 & 47.0 & 89.7 & 66.7 & 35.9 & 90.2 & 48.9 & 57.5 & 0.0 & 39.8 & 56.0 & 56.6\\
		BAPA \cite{liu2021bapa} & 94.4 & 61.0 & 88.0 & 26.8 & 39.9 & 38.3 & 46.1 & 55.3 & 87.8 & 46.1 & 89.4 & 68.8 & 40.0 & 90.2 & 60.4 & 59.0 & 0.0 & 45.1 & 54.2 & 57.4\\
		ProDA \cite{PanZhang2021PrototypicalPL} & 87.8 & 56.0 & 79.7 & 46.3 & 44.8 & 45.6 & 53.5 & 53.5 & 88.6 & 45.2 & 82.1 & 70.7 & 39.2 & 88.8 & 45.5 & 59.4 & 1.0 & 48.9 & 56.4 & 57.5\\
        CaCo \cite{huang2022category} & 93.8 & 64.1 & 85.7 & 43.7 & 42.2 & 46.1 & 50.1 & 54.0 & 88.7 & 47.0 & 86.5 & 68.1 & 2.9 & 88.0 & 43.4 & 60.1 & 31.5 & 46.1 & 60.9 & 58.0\\
        \hline
		DAFormer \cite{hoyer2022daformer} & 95.7 & 70.2 & 89.4 & 53.5 & 48.1 & 49.6 & 55.8 & 59.4 & 89.9 & 47.9 & 92.5 & 72.2 & 44.7 & 92.3 & 74.5 & 78.2 & 65.1 & 55.9 & 61.8 & 68.3\\
        CAMix \cite{zhou2022context} & 96.0 & 73.1 & 89.5 & 53.9 & 50.8 & 51.7 & 58.7 & 64.9 & 90.0 & 51.2 & 92.2 & 71.8 & 44.0 & 92.8 & 78.7 & 82.3 & 70.9 & 54.1 & 64.3 & 70.0\\
        \rowcolor{lightgray} 
		DAFormer \cite{hoyer2022daformer} + PiPa  & 96.1 & 72.0 & 90.3 & 56.6 & 52.0 & 55.1 & 61.8 & 63.7 & 90.8 & \textbf{52.6} & 93.6 & 74.3 & 43.6 & 93.5 & 78.4 & 84.2 & 77.3 & 59.9 & 66.7 & 71.7\\
        \hline
		HRDA \cite{hoyer2022hrda} & 96.4 & 74.4 & 91.0 & 61.6 & 51.5 & 57.1 & 63.9 & 69.3 & 91.3 & 48.4 & 94.2 & 79.0 & 52.9 & 93.9 & 84.1 & 85.7 & 75.9 & \textbf{63.9} & \textbf{67.5} & 73.8\\
		\rowcolor{lightgray} 
		HRDA \cite{hoyer2022hrda} + PiPa & \textbf{96.8} & \textbf{76.3} & \textbf{91.6} & \textbf{63.0} & \textbf{57.7} & \textbf{60.0} & \textbf{65.4} & \textbf{72.6} & \textbf{91.7} & 51.8 & \textbf{94.8} & \textbf{79.7} & \textbf{56.4} & \textbf{94.4} & \textbf{85.9} & \textbf{88.4} & \textbf{78.9} & 63.5 & 67.2 & \textbf{75.6}\\
		\shline
	\end{tabular}
	}
\end{table*}

\subsection{Implementation Details} 
\noindent\textbf{Datasets.} 
We evaluate the proposed method on two unsupervised domain adaptation settings, \ie, GTA $\rightarrow$ Cityscapes and SYNTHIA $\rightarrow$ Cityscapes, following common UDA protocols \cite{WilhelmTranheden2020DACSDA, wang2021domain, araslanov2021self, hoyer2022daformer, hoyer2022hrda}. 
Particularly, the GTA dataset is the synthetic dataset collected from the video game "Grand Theft Auto V", yielding 24,966 city scene images and 19 classes. SYNTHIA is a synthetic urban scene dataset with 9,400 training images and 16 classes. GTA and SYNTHIA are utilized as source domain datasets. The target dataset Cityscapes, collected from the real-world street-view images, contains 2,975 unlabeled training images and 500 images for testing.

\noindent\textbf{Structure Details.} 
Following recent SOTA UDA setting \cite{hoyer2022daformer,zhou2022context, BinhuiXie2022SePiCoSP}, our network consists of a SegFormer MiT-B5 backbone~\cite{xie2021segformer,hoyer2022daformer} pretrained on ImageNet-1k~\cite{deng2009imagenet} and several MLP-based heads, \ie, $h_{\text {cls }}$, $h_{\text {pixel }}$ and $h_{\text {patch }}$, which contains two fully-connected (fc) layers and ReLU activation between two fc layers. 
Note that the self-supervised projection heads $h_{\text {pixel }}$ and $h_{\text {patch }}$ are only applied at training time and are removed during inference, which does not introduce extra computational costs in deployment. 

\noindent\textbf{Implementation details.}
We train the network with batch size 2 for 60k iterations with a single NVIDIA RTX 6000 GPU. 
We adopt AdamW \cite{loshchilov2017decoupled} as the optimizer, a learning rate of $6 \times 10^{-5}$, a linear learning rate warmup of $1.5 \mathrm{k}$ iterations and the weight decay of 0.01. 
Following \cite{zhou2022context, BinhuiXie2022SePiCoSP}, the input image is resized to 1280 $\times$ 720 for GTA and 1280 $\times$ 760 for SYNTHIA, with a random crop size of 640 $\times$ 640. For the patch-wise contrast, we randomly resize the input images by a ratio between 0.5 and 2, and then randomly crop two patches of the size $720 \times 720$ from the resized image and ensure the Intersection-over-Union(IoU) value of the two patches between 0.1 and 1. 
We utilize the same data augmentation \eg, color jitter, Gaussian blur and ClassMix \cite{olsson2021classmix} and empirically set pseudo labels threshold $0.968$ following \cite{WilhelmTranheden2020DACSDA}. The exponential moving average parameter of the teacher network is $0.999$. The hyperparameters of the loss function are chosen empirically $\alpha = \beta = 0.1$. 
We evaluate our methods using the mean intersection-over-union (mIoU). We report our results for 19 classes on GTA $\rightarrow$ Cityscapes and both 13 and 16 classes on SYNTHIA $\rightarrow$ Cityscapes.

\noindent\textbf{Reproducibility.} The code is based on Pytorch \cite{paszke2017automatic}. We will make our code open-source for reproducing all results.


\begin{table*}[!t]
	\centering
	\caption{Quantitative comparison with previous UDA methods on SYNTHIA $\rightarrow$ Cityscapes. We present pre-class IoU, mIoU and mIoU*. mIoU and mIoU* are averaged over 16 and 13 categories, respectively. The best accuracy in every column is in \textbf{bold}.}
	\label{table:syncity}
	\resizebox{\linewidth}{!}{
	\begin{tabular}{c|cccccccccccccccc|c|c}
		\shline
		Method & Road & SW & Build & Wall* & Fence* & Pole* & TL & TS & Veg. & Sky & PR & Rider & Car & Bus & Motor & Bike & mIoU* & mIoU\\
		\shline
		\hline
		MaxSquare~\cite{chen2019domain} & 77.4 & 34.0 & 78.7  & 5.6 & 0.2 & 27.7 & 5.8 & 9.8 & 80.7 & 83.2 & 58.5 & 20.5 & 74.1 & 32.1 & 11.0 & 29.9 & 45.8 & 39.3 \\
		SIBAN~\cite{luo2019significance} & 82.5 & 24.0 & 79.4 & $-$ & $-$ & $-$ & 16.5 & 12.7 & 79.2 & 82.8 & 58.3 & 18.0 & 79.3 & 25.3 & 17.6 & 25.9 & 46.3 & $-$ \\
    	PatchAlign~\cite{tsai2019domain} & 82.4 & 38.0 & 78.6 & 8.7 & 0.6 & 26.0 & 3.9 & 11.1 & 75.5 & 84.6 & 53.5 & 21.6 & 71.4 & 32.6 & 19.3 & 31.7 & 46.5 & 40.0 \\
		AdaptSegNet~\cite{tsai2018learning} & 84.3 & 42.7 & 77.5 & $-$ & $-$ & $-$ & 4.7 & 7.0 & 77.9 & 82.5 & 54.3 & 21.0 & 72.3 & 32.2 & 18.9 & 32.3 & 46.7 & $-$ \\
		CLAN~\cite{luo2019taking} & 81.3 & 37.0 & 80.1 & $-$ & $-$ & $-$ & 16.1 & 13.7 & 78.2 & 81.5 & 53.4 & 21.2 & 73.0 & 32.9 & 22.6 & 30.7 & 47.8 & $-$  \\
		SP-Adv~\cite{SHAN2020125} & 84.8 & 	35.8 & 	78.6 & $-$ & $-$ & $-$ &	6.2	& 15.6 & 	80.5 &	82.0 &	66.5 & 	22.7 & 	74.3 &	34.1	 & 19.2	 & 27.3 & 	48.3 &  $-$ \\
        AdvEnt \cite{vu2019advent} & 85.6 & 42.2 & 79.7 & 8.7 & 0.4 & 25.9 & 5.4 & 8.1 & 80.4 & 84.1 & 57.9 & 23.8 & 73.3 & 36.4 & 14.2 & 33.0 & 48.0 & 41.2 \\
		ASA~\cite{zhou2020affinity} & \textbf{91.2} & 48.5 & 80.4 & 3.7 & 0.3 & 21.7 & 5.5 & 5.2 & 79.5 & 83.6 & 56.4 & 21.0 & 80.3 & 36.2 & 20.0 & 32.9 & 49.3 & 41.7 \\
		CBST \cite{YangZou2018UnsupervisedDA} & 68.0 & 29.9 & 76.3 & 10.8 & 1.4 & 33.9 & 22.8 & 29.5 & 77.6 & 78.3 & 60.6 & 28.3 & 81.6 & 23.5 & 18.8 & 39.8 & 48.9 & 42.6 \\
		DADA~\cite{vu2019dada} & 89.2 & 44.8 & 81.4 & 6.8 & 0.3 & 26.2 & 8.6 & 11.1 & 81.8 & 84.0 & 54.7 & 19.3 & 79.7 & 40.7 & 14.0 & 38.8 & 49.8 & 42.6 \\
		MRNet~\cite{zheng2019unsupervised} & 82.0 & 36.5 & 80.4 & 4.2 & 0.4 & 33.7 & 18.0 & 13.4 & 81.1 & 80.8 & 61.3 & 21.7 & 84.4 & 32.4 & 14.8 & 45.7 & 50.2 & 43.2 \\
		MRKLD \cite{zou2019confidence} & 67.7 & 32.2 & 73.9 & 10.7 & 1.6 & 37.4 & 22.2 & 31.2 & 80.8 & 80.5 & 60.8 & 29.1 & 82.8 & 25.0 & 19.4 & 45.3 & 50.1 & 43.8 \\
		CCM~\cite{li2020content} & 79.6 & 36.4 & 80.6 & 13.3 & 0.3 & 25.5 & 22.4 & 14.9 & 81.8 & 77.4 & 56.8 & 25.9 & 80.7 & 45.3 & 29.9 &  52.0 & 52.9 & 45.2 \\
		Uncertainty~\cite{zheng2021rectifying} & 87.6 & 41.9 & 83.1 & 14.7 & 1.7 & 36.2 & 31.3 & 19.9 & 81.6 & 80.6 & 63.0 & 21.8 & 86.2 & 40.7 & 23.6 & 53.1 & 54.9 & 47.9 \\
		BL~\cite{li2019bidirectional} & 86.0 & 46.7 & 80.3 & $-$ & $-$ & $-$ & 14.1 & 11.6 & 79.2 & 81.3 & 54.1 & 27.9 & 73.7 & 42.2 & 25.7 & 45.3 & 51.4 & $-$ \\
		DT~\cite{wang2020differential} & 83.0 & 44.0 & 80.3 & $-$ & $-$ & $-$ & 17.1 & 15.8 & 80.5 & 81.8 & 59.9 & 33.1 & 70.2 & 37.3 & 28.5 & 45.8 & 52.1 &  $-$ \\
		APODA~\cite{yang2020adversarial} & 86.4 & 41.3 & 79.3 & $-$ & $-$ & $-$ & 22.6 & 17.3 & 80.3 & 81.6 & 56.9 & 21.0 & 84.1 & 49.1 & 24.6 & 45.7 & 53.1 & $-$  \\
	    Adaboost~\cite{zheng2022adaptive} & 85.6 & 43.9 & 83.9 & 19.2 & 1.7 & 38.0 & 37.9 & 19.6 & 85.5 & 88.4 & 64.1 & 25.7 & 86.6 & 43.9 & 31.2 & 51.3 & 57.5 & 50.4 \\
		\hline
		DAFormer \cite{hoyer2022daformer} & 84.5 & 40.7 & 88.4 & 41.5 & 6.5 & 50.0 & 55.0 & 54.6 & 86.0 & 89.8 & 73.2 & 48.2 & 87.2 & 53.2 & 53.9 & 61.7 & 67.4 & 60.9 \\
        CAMix \cite{zhou2022context} & 87.4 & 47.5 & 88.8 & $-$ & $-$ & $-$ & 55.2 & 55.4 & 87.0 & 91.7 & 72.0 & 49.3 & 86.9 & 57.0 & 57.5 & 63.6 & 69.2 & $-$ \\
		\rowcolor{lightgray} 
		DAFormer \cite{hoyer2022daformer} + PiPa & 87.9 & 48.9 & 88.7 & 45.1 & 4.5 & 53.1 & 59.1 & 58.8 & 87.8 & 92.2 & 75.7 & 49.6 & 88.8 & 53.5 & 58.0 & 62.8 & 70.1 & 63.4 \\
		\hline
        HRDA \cite{hoyer2022hrda} & 85.2 & 47.7 & 88.8 & 49.5 & 4.8 & 57.2 & 65.7 & 60.9 & 85.3 & 92.9 & 79.4 & 52.8 & 89.0 & 64.7 & 63.9 & 64.9 & 72.4 & 65.8 \\
		\rowcolor{lightgray} 
        HRDA \cite{hoyer2022hrda} + PiPa & 88.6 & \textbf{50.1} & \textbf{90.0} & \textbf{53.8} & \textbf{7.7} & \textbf{58.1} & \textbf{67.2} & \textbf{63.1} & \textbf{88.5} & \textbf{94.5} & \textbf{79.7} & \textbf{57.6} & \textbf{90.8} & \textbf{70.2} & \textbf{65.1} & \textbf{66.9} & \textbf{74.8} & \textbf{68.2} \\

		\shline
	\end{tabular}
	}
\end{table*}

\begin{figure*}[!t]
  \centering
    \includegraphics[width=0.95\linewidth]{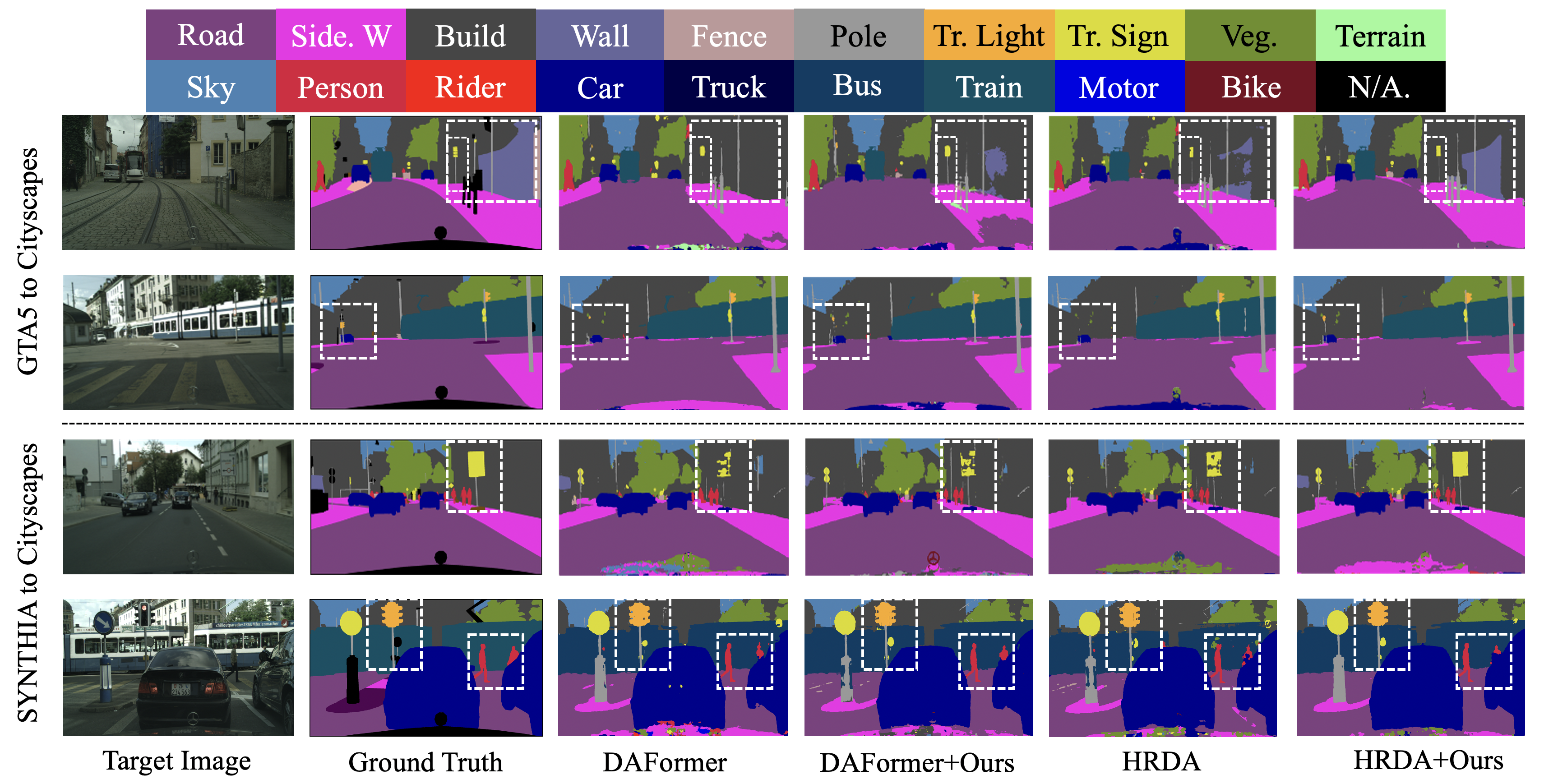}
    \vspace{-.1in}
    \caption{Qualitative results on GTA $\rightarrow$ Cityscapes and SYNTHIA $\rightarrow$ Cityscapes. From left to right: Target Image, Ground Truth, the visual results predicted by DAFormer, DAFormer + Ours (PiPa), HRDA, HRDA + Ours (PiPa). We deploy the white dash boxes to highlight different prediction parts. }
    \label{fig3}
\end{figure*}

\subsection{Comparisons with State-of-the-art Methods}
We compare PiPa with several competitive UDA methods on GTA $\rightarrow$ Cityscapes and SYNTHIA $\rightarrow$ Cityscapes, respectively. The quantitative comparisons are shown in Table \ref{table:gtacity} and Table \ref{table:syncity}, respectively. We also show the visual difference between the proposed method and the other two strong Transformer-based methods \cite{hoyer2022daformer, hoyer2022hrda} in Figure~\ref{fig3}. 

\noindent\textbf{GTA $\rightarrow$ Cityscapes.}  First, we show our results on GTA $\rightarrow$ Cityscapes in Table \ref{table:gtacity} and highlight the best results in bold. It could be observed that 
Transformer-based methods significantly surpass the CNN-based SOAT by a large margin since DAFormer \cite{hoyer2022daformer}. Generally, our PiPa yields a significant improvement over the previous strongest transformer-based models DAFormer\cite{hoyer2022daformer} and HRDA\cite{hoyer2022hrda}. Particularly, PiPa achieves 71.7 mIoU, which outperforms DAFormer by a considerable margin of +3.4 mIoU. Additionally, when applying PiPa to HRDA, which is a strong baseline that adopts high-resolution crops, we increase +1.8 mIoU and achieve the state-of-the-art performance of 75.6 mIoU, verifying the effectiveness of the proposed method that introduces a unified and multi-grained self-supervised learning algorithm in UDA task. Furthermore, PiPa achieves leading IoU of almost all classes on GTA $\rightarrow$ Cityscapes, including several small-scale objectives such as Fence, Pole, Wall and Training Sign. Particularly, we increase the IoU of the Fence by +6.2 from 51.5 to 57.7 IoU. The IoU performance of PiPa verifies our motivation that the exploration of the inherent structures of intra-domain images indeed helps category recognition, especially for challenging small objectives.

\noindent\textbf{SYNTHIA $\rightarrow$ Cityscapes.}
As revealed in Table \ref{table:syncity}, PiPa also achieves remarkable mIoU and mIoU* (13 most common categories) performance on SYNTHIA $\rightarrow$ Cityscapes, increasing +2.5 and +2.4 mIoU compared with DAFormer \cite{hoyer2022daformer} and HRDA \cite{hoyer2022hrda}, respectively. It is noticeable that our PiPa remains competitive in segmenting each individual class including small-scale objectives and yields the best IoU score in almost all categories.

\noindent\textbf{Qualitative results.}
In Figure~\ref{fig3}, we visualize the segmentation results and the comparison with previous strong methods DAFormer \cite{hoyer2022daformer}, HRDA \cite{hoyer2022hrda}, and the ground truth on both GTA $\rightarrow$ Cityscapes and SYNTHIA $\rightarrow$ Cityscapes benchmarks. The results highlighted by white dash boxes show that PiPa is capable of segmenting minor categories such as `wall', `traffic sign' and `traffic light'. It is also noticeable that PiPa predicts smoother edges between different categories, \eg, `person' in the fourth row of Figure~\ref{fig3}. We think it is because the proposed method explicitly encourages patch-wise consistency against different contexts, which facilitates the prediction robustness on edges.

\subsection{Ablation Studies and Further Analysis}

\begin{figure}[t]
  \centering
  \includegraphics[width=0.95\linewidth]{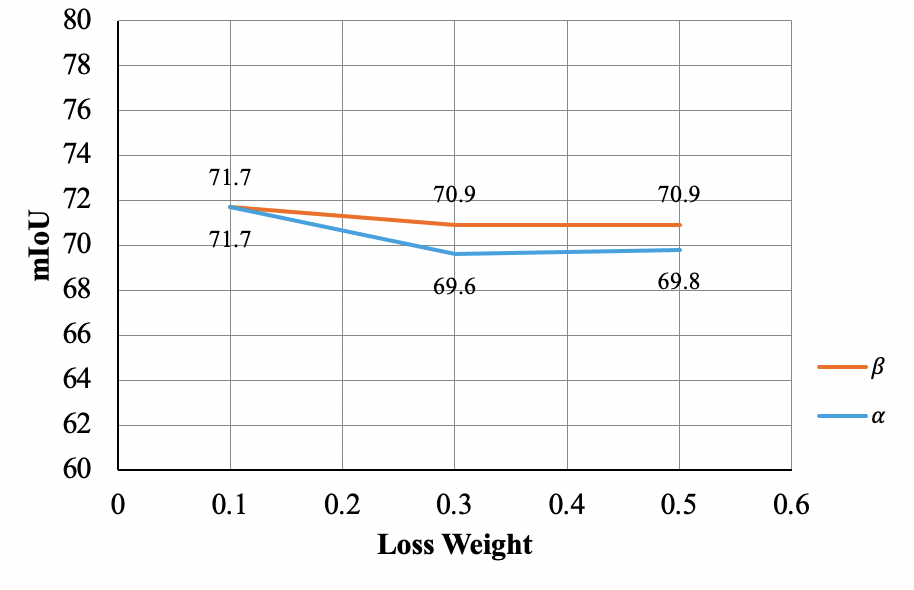}
  \vspace{-.2in}
  \caption{Ablation study on Loss Weights $\alpha$ and $\beta$}
  \label{fig4}
\end{figure}

\noindent\textbf{Effect of Pixel-wise Contrast and Patch-wise Contrast.} 
We evaluate the effectiveness of the two primary components, \ie, Pixel-wise Contrast and Patch-wise Contrast in the proposed PiPa and investigate how the combination of two contrasts contributes to the final performance on GTA $\rightarrow$ Cityscapes. 
For a fair comparison, we apply the same experimental settings and hyperparameters. 
The baseline model is based on DAFormer, which yields a competitive mIoU of 68.4. 
As shown in the Table \ref{table:ablation}, we could observe: 
(1) Both Patch Contrast and Pixel Contrast individually could lead to +1.4 mIoU and +2.3 mIoU improvement respectively, verifying the effectiveness of exploring the inherent contextual knowledge. 
(2) The two kinds of contrasts are complementary to each other. The proposed method successfully mines the multi-level knowledge by combining the two kinds of contrast. When applying both losses, our PiPa further improves the network performance to 71.7 mIoU, surpassing the model that deploys only one kind of contrast by a clear margin. 

\begin{table}[t]
    \centering
    \caption{Ablation study on the effect of Pixel-wise Contrast and Patch-wise Contrast on GTA $\rightarrow$ Cityscapes.} 
    \label{table:ablation}
    \begin{tabular}{l|c|c|c|c}
    \shline
    Method & $\mathcal{L}_{\text{Pixel}}$ & $\mathcal{L}_{\text{Patch}}$ & mIoU & $\Delta$mIoU\\
    \hline 
    DAFormer\cite{hoyer2022daformer} & & & $68.4$ & $-$ \\
    Patch Contrast &  & $\checkmark$ & $69.8$ & +1.4 \\
    Pixel Contrast & $\checkmark$ & & $70.7$ & +2.3 \\
    PiPa & $\checkmark$ & $\checkmark$ & $71.7$ & +3.3 \\
    \shline
    \end{tabular}
\end{table}

\noindent\textbf{Effect of the loss weight.}
We conduct loss weight sensitivity analysis on GTA $\rightarrow$ Cityscapes. 
Specifically, we change the weights $\alpha$ and $\beta$ of the two kinds of contrasts in Eq~\ref{eq:6}, respectively.
As shown in Figure \ref{fig4}, we can observe that both pixel-wise and patch-wise contrast are not sensitive to the relative weight. PiPa keeps outperforming the competitive DAFormer baseline of 68.3 mIoU in all compositions of loss weights. 
When applying the proposed method to a unseen environment, $\alpha=0.1, \beta=0.1$ can be a good initial weight to start.

\noindent\textbf{Effect of the patch crop size.}
For the patch contrast, the size of the patch also affects the number of negative pixels and training difficulty. As shown in Table \ref{table:crop}, we gradually increase the patch size. We observe that larger patch generally obtain better performance since it contains more diverse contexts and is close to the test size. The best performance is 71.7 mIoU when the cropped patch size is $720\times720$.
It is also worth noting that if the patch size is too large (like 960), the overlapping area can be larger than the non-overlapping area, which also may compromise the training.  

\begin{table}[t]
    \centering
    \caption{Effect of the patch crop size on GTA $\rightarrow$ Cityscapes.}
    \label{table:crop} \small
    \begin{tabular}{c|c}
    \shline Crop Size & $\mathrm{mIoU}$ \\
    \hline 
    $480 \times 480$  & $70.4$ \\
    $600 \times 600$  & $71.0$ \\
    $720 \times 720$  & $71.7$ \\
    $900 \times 900$  & $70.9$ \\
    \shline
    \end{tabular}
\end{table}

\section{Conclusion}
In this work, we focus on the exploration of intra-domain knowledge, such as context correlation inside an image for the semantic segmentation domain adaptation. We target to learn a feature space that enables discriminative pixel-wise features and the robust feature learning of the overlapping patch against variant contexts. 
To this end, we propose PiPa, a unified pixel- and patch-wise self-supervised learning framework, which introduces pixel-level and patch-level contrast learning to UDA. PiPa encourages the model to mine the inherent contextual feature, which is domain invariant.
Experiments show that PiPa outperforms the state-of-the-art approaches and yields a competitive 75.6 mIoU on GTA$\rightarrow$Cityscapes and 67.4 mIoU on Synthia$\rightarrow$Cityscapes. 
Since PiPa does not introduce extra parameters or annotations, it can be combined with other existing methods to further facilitate the intra-domain knowledge learning. In the future, we will continue to study the proposed PiPa on relevant tasks, such as domain adaptive video segmentation and open-set adaptation \emph{etc.}

{\small
\bibliographystyle{ieee_fullname}
\bibliography{egbib}
}

\end{document}